\documentclass[11pt]{article}

\usepackage[preprint]{acl}

\usepackage{times}
\usepackage{latexsym}

\usepackage[T1]{fontenc}

\usepackage[utf8]{inputenc}

\usepackage{microtype}

\usepackage{inconsolata}

\usepackage{graphicx}

\usepackage{amsmath}
\usepackage{amssymb}
\usepackage{listings}

\title{Activation-Guided Neuron Intervention to Induce Alzheimer's-Related Computational Language Phenotypes in a Large Language Model}

\author{
\textbf{Rui He\textsuperscript{1,*}},
\textbf{Ercong Nie\textsuperscript{2}},
\textbf{Hong Jiang\textsuperscript{3}},
\textbf{Iris E. Sommer\textsuperscript{4}},
\textbf{Philipp Homan\textsuperscript{5,6}},
\textbf{Wolfram Hinzen\textsuperscript{1,7}}
\\[0.2em]
\parbox{0.9\textwidth}{
\centering
\normalsize
\textsuperscript{1}Grammar and Cognition Lab, Department of Translation \& Language Sciences, Universitat Pompeu Fabra, Barcelona, Spain.\\
\textsuperscript{2}School of Foreign Languages, Shanghai Jiao Tong University,
Shanghai, China.\\
\textsuperscript{3}Zhuhai People's Hospital (The Affiliated Hospital of Beijing Institute of Technology, Zhuhai Clinical Medical College of Jinan University), Zhuhai, China.\\
\textsuperscript{4}Center for Clinical Neuroscience and Cognition and Department of Psychiatry, University of Groningen, University Medical Center Groningen,
Groningen, the Netherlands.\\
\textsuperscript{5}Department of Adult Psychiatry and Psychotherapy, University Hospital of Psychiatry Zurich, University of Zurich, Zurich, Switzerland.\\
\textsuperscript{6}Neuroscience Center Zurich, University of Zurich and ETH Zurich, Zurich, Switzerland.\\
\textsuperscript{7}Institució Catalana de Recerca i Estudis Avançats, Barcelona, Spain.\\[0.15em]
\textbf{\textsuperscript{*}Correspondence:}
\href{mailto:rui.he@upf.edu}{rui.he@upf.edu}
}
}

\begin{document}

\maketitle

\begin{abstract}
Changes in spontaneous speech provide an early signal of cognitive dysfunction in Alzheimer’s disease (AD) that large language models (LLMs) can detect. However, detection alone cannot establish whether the underlying model representations contribute functionally to behavior. We introduce an activation-guided intervention framework using Qwen3-8B. The framework identifies feed-forward neurons with higher activation rates for AD than control transcripts and modulates their output contributions during generation by scaling the corresponding down-projection weights. This yielded nine edited variants differing in intervention direction, magnitude, and scope. The original and edited models completed the same 12-turn neuropsychological battery, assessed through blinded human ratings and computational linguistic measures. Amplifying AD-associated neurons produced graded impairments in story recall, verbal fluency, working memory, procedural discourse, scene construction, and coreference resolution. Attenuation largely preserved performance and selectively improved several outcomes. Amplification also reduced lexical surprisal, idea density, syntactic complexity, and discourse quantity, broadly paralleling changes reported in human AD speech. These findings show that neurons identified solely from clinical language differences can influence behavior across multiple cognitive domains, providing proof of concept for an AD-related computational phenotype and a controlled framework for experimentally examining links between language and broader cognitive dysfunction.\footnote{\url{https://github.com/RuiHe1999/ad-neuron-intervention}}

\end{abstract}

\section{Introduction}
Changes in language processing are among the most accessible manifestations of cognitive decline in Alzheimer’s disease (AD) \citep{shankar_systematic_2025}. Computational analyses of speech and language can detect subtle alterations at early stages of the AD continuum, including subjective cognitive decline and amyloid-positive cognitively unimpaired individuals, with more pronounced changes in mild cognitive impairment and clinically manifest dementia \citep{shankar_systematic_2025,fristed_leveraging_2022,he_automated_2023,van_den_berg_digital_2024}. Unlike many biological markers, speech and language can be sampled repeatedly, remotely, and at relatively low cost, making them promising digital markers \citep{robin_evaluation_2020}.

Decades of computational efforts began with major reliance on predefined acoustic and linguistic features, including speech timing, lexical diversity, syntactic complexity, lexical surprisal, and information density \cite{fraser_linguistic_2015, luz_alzheimers_2020}. Across studies, AD speech shows greater hesitation and word-finding difficulty, lower lexical diversity and information content, greater reliance on frequent words, and reduced syntactic complexity \citep{boschi_connected_2017,slegers_connected_2018}. Later work used distributed representations and language models to find altered semantic similarity, higher perplexity, and weaker stimulus-response alignment \citep{colla_semantic_2022,gkoumas-etal-2023-digital,jiang_structure_2025}, while more recent studies have learned AD-related patterns directly from patient language through fine-tuning, representation learning, and prompting \citep{balagopalan_bert_2020,farzana_domain_2024,heitz_linguistic_2025}.

These advances extend the role of computational models beyond prediction alone. Beyond their use as predictors, pretrained language models have also become tools for examining how linguistic information is organized in internal representations that can support AD prediction \citep{balagopalan_bert_2020,farzana_domain_2024}. Probing and localization studies have associated particular layers, components, and neurons with syntactic, semantic, factual, and stylistic properties \citep{jawahar_what_2019,dai-etal-2022-knowledge,lai_style-specific_2024}. Such associations, however, do not by themselves establish whether the identified components contribute functionally to model behavior. Controlled perturbation provides a stronger test by examining how behavior changes when candidate components are disrupted. Previous studies have used lesioning, component ablation, and broader parameter perturbation to relate predefined layers or computational components to linguistic capacities and clinically interpretable behavioral changes \citep{li_gpt-d_2022, wang_component-level_2026, yang_lesioned_2026}.

Intervening on the in silico LLM neurons offers a more selective form of this approach. Rather than disrupting an entire layer, matrix, or architectural component, it modifies the contribution of a restricted set of internal units while preserving the remaining model architecture and non-target parameters \citep{dai-etal-2022-knowledge,mueller-etal-2022-causal,xiao_neuron-based_2025}. The resulting changes offer a controlled setting to test whether identified internal representational differences contribute to model behavior rather than merely correlate with it \citep{mueller-etal-2022-causal,lai_style-specific_2024}. The purpose of such editing is not simply to reproduce surface-level behavior associated with AD. Rather, we aim to develop an experimentally manipulable model of how AD-related language differences may arise from changes in internal representations. Such a model moves from prediction toward mechanism by linking selective internal perturbations to their downstream behavioral consequences. Whether neuron-level intervention can yield a coherent and clinically meaningful AD-related phenotype, however, remains largely unexplored.

We therefore ask: \textbf{Q1:} Can neurons associated with AD-control language differences be identified and selectively modulated? \textbf{Q2:} Does increasing their contribution induce systematic changes across cognitive domains that parallel impairments observed in human AD? \textbf{Q3:} Do these effects vary with intervention direction, magnitude, and scope?

To address these questions, we identify AD-associated neurons from activation differences elicited by transcripts from individuals with AD and cognitively healthy older adults. We then amplify or attenuate their output contributions by scaling the corresponding down-projection weights at different magnitudes and intervention scopes. The original and edited models are evaluated using an identical multi-turn neuropsychological battery spanning multiple cognitive domains. We use the term \emph{AD-related computational (language) phenotype} to denote the reproducible behavioral profile induced by editing these LLM units. This term refers only to the behavioral pattern of the LLM and does not imply that the model reproduces AD pathology.

Our contributions are threefold: \textbf{(i)} an activation-guided framework for targeted editing of AD-associated neurons; \textbf{(ii)} systematic evaluation across editing directions, magnitudes, and scopes; and \textbf{(iii)} evidence that entirely language-grounded intervention affects multiple neuropsychological domains, revealing the functional entanglement of language with broader cognition.

\section{Related work}

\paragraph{Controlled model perturbation in clinical language.}
Controlled perturbation of language models has emerged as a promising approach for examining whether disruptions to internal computation can produce clinically interpretable linguistic behavior. GPT-D provided an early example by degrading attention parameters in GPT-2, inducing dementia-related linguistic anomalies and using perplexity to distinguish language produced by individuals with dementia from that of cognitively healthy individuals \citep{li_gpt-d_2022}. Subsequent work extended this approach through attention-head ablation and examined how model scale affects susceptibility to induced dementia-related degradation \citep{li_too_2024}. Recent aphasia research has similarly used expert lesioning, component-level ablation, and graded parameter disruption to induce language impairments and compare the resulting profiles with clinical aphasia phenotypes \citep{wang_emergent_2025,wang_component-level_2026,roll_artificial_2026,yang_lesioned_2026}. Together, these studies establish language models as controlled systems for linking internal disruption to clinically relevant phenotypes. Our intervention differs in both its granularity and empirical basis: rather than perturbing components such as layers or parameter matrices selected a priori, we identify individual neurons from AD-control activation differences and modulate those same units during generation.

\paragraph{Neuron-level intervention.}
Neuron-level intervention advances mechanistic interpretability by testing whether localized information contributes functionally to model behavior. Early work identified neurons encoding linguistic properties and showed that manipulating their activations could alter model outputs \citep{bau_identifying_2018}. Transformer feed-forward layers were subsequently characterized as key-value memory systems in which input patterns retrieve stored representations \citep{geva_transformer_2021}. Building on this account, specific factual associations were localized to restricted sets of feed-forward neurons \citep{dai-etal-2022-knowledge}. Localization alone, however, does not establish that the identified representation contributes functionally to model output. Sparse masking and causal editing therefore introduced intervention-based methods for testing whether modifying localized internal representations produces corresponding behavioral changes \citep{de_cao_sparse_2022, meng_locating_2022}. More recent work has used activation contrasts for controllable properties of generated language, for example, to identify and modulate neurons associated with style for altering stylistic expression \citep{lai_style-specific_2024} and language selection to reduce unintended language switching \citep{nie_mechanistic_2025}. These findings demonstrate that behaviorally defined activation contrasts can serve both to localize candidate units and to test their functional contribution through targeted intervention. We extend this locate-and-intervene framework to clinical language by deriving intervention targets from AD-related activation differences and evaluating their effects across neuropsychological tasks.

\section{Methods}
\begin{figure*}[t]
  \includegraphics[width=\linewidth]{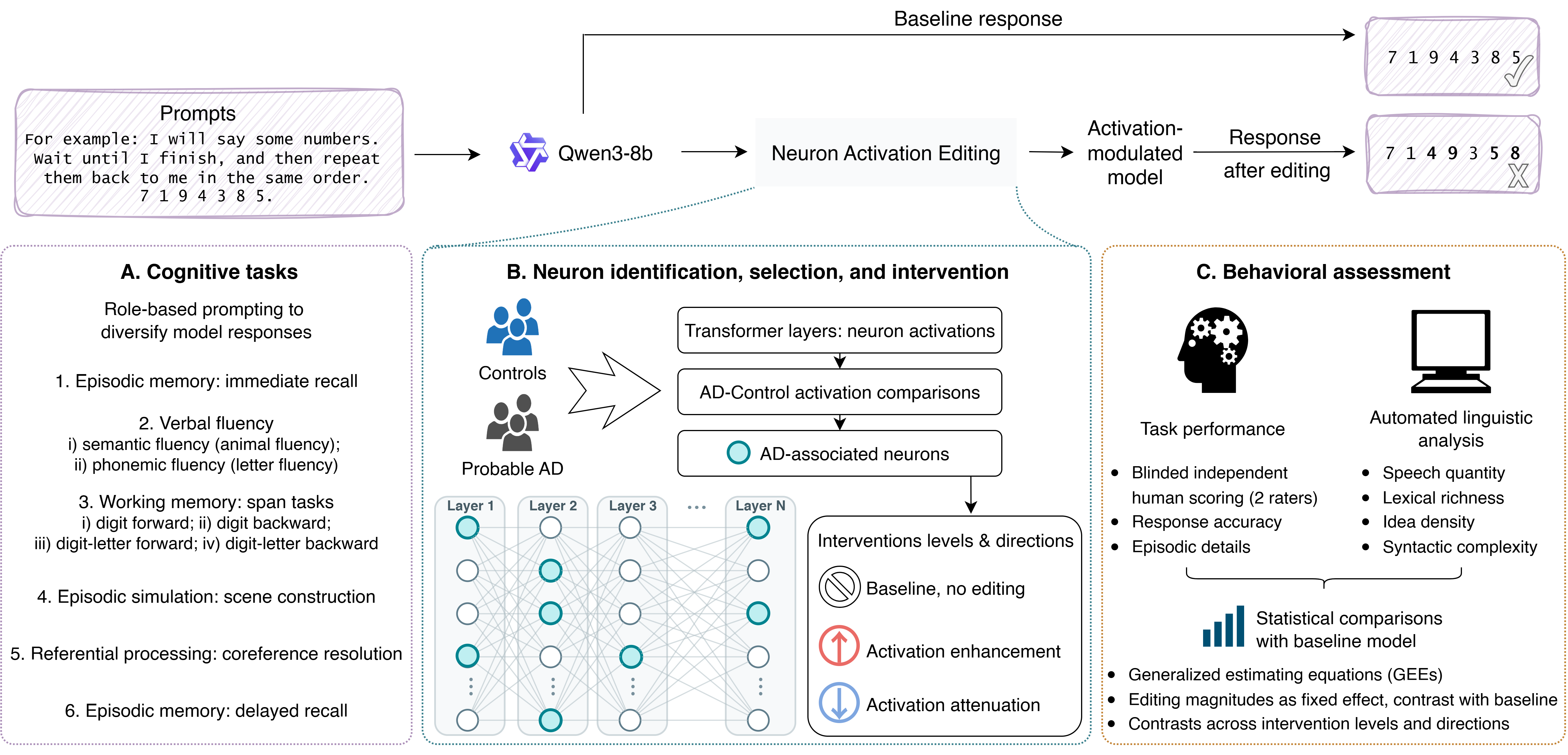} 
  \caption {
  \textbf{Overview of the neuron-level intervention framework.}
  \textbf{(A)} An adapted neuropsychological battery was administered to Qwen3-8B using role-based prompts.
  \textbf{(B)} Neurons with significantly higher activation rates for transcripts from individuals with probable Alzheimer's disease (AD) than for those from cognitively healthy older adults were identified and selectively modulated during generation.
  \textbf{(C)} Intervention effects were assessed using task-performance measures and automated linguistic analyses, with each intervention condition compared with the unedited model using generalized estimating equations.
  }
  \label{fig:workflow}
\end{figure*}

Figure~\ref{fig:workflow} summarizes the framework. We first identified feed-forward neurons with higher activation rates for AD than control transcripts, scaled their down-projection weights to amplify or attenuate their output contributions, and evaluated the edited models on language-based tasks spanning multiple cognitive domains (Section~\ref{sec:experiment}).

\subsection{Identify AD-associated neurons}
\subsubsection{Clinical language data}
We used picture description recordings from the ADReSSo dataset to identify AD-associated neurons \citep{luz_detecting_2021}. ADReSSo includes participants with probable AD and cognitively healthy older adults matched for age and sex using propensity scores, thereby improving group comparability. Audio recordings were linked to their corresponding DementiaBank transcripts. A matching transcript was available for all but one recording, which we manually transcribed. Only participant speech was retained, yielding one transcript per participant for activation extraction. Dataset and preprocessing details are provided in Appendix~\ref{sec:appendix_adresso}.

\subsubsection{LLM neurons and their activations}
We define a neuron as a scalar unit in the intermediate representation of the feed-forward network (FFN) at a specific Transformer layer \citep{bau_identifying_2018}. Model parameters remained frozen, and activations were extracted in evaluation mode without gradient computation. 

Let $\mathbf{h}^{l}_{t}$ denote the hidden representation of token $t$ entering the FFN at layer $l$. The post-activation gate vector is

\begin{equation}
\mathbf{a}^{l}_{t}=
\operatorname{SiLU}
\left(
\mathbf{W}^{l}_{\mathrm{gate}}
\mathbf{h}^{l}_{t}
\right),
\end{equation}

where $\mathbf{W}^{l}{\mathrm{gate}}$ is the gate-projection matrix and $a^{l}_{t,i}$ denotes the activation of neuron $i$. 

Following \citet{lai_style-specific_2024}, neuron $i$ was considered active for token $t$ when $a^{l}_{t,i}>0$. This definition was applied to every intermediate neuron in every Transformer layer. Each transcript was tokenized without padding. To obtain a transcript-level measure of neuronal activity, we calculated the proportion of tokens for which each neuron was active. For a transcript $x$ containing $T_x$ tokens, the activation rate of neuron $i$ in layer $l$ was defined as

\begin{equation}
r^{l}_{i}(x)=
\frac{1}{T_x}
\sum_{t=1}^{T_x}
\mathbb{I}
\left[
a^{l}_{t,i}>0
\right],
\label{eq:activation_rate}
\end{equation}

where $\mathbb{I}$ is the indicator function. The resulting activation rate ranges from zero to one. A value close to one indicates that the neuron was active for most tokens in the transcript, whereas a value close to zero indicates that it was rarely active. This aggregation yielded one length-normalized activation rate for each participant, layer, and neuron.

\subsubsection{Neuron selection}
Participant-level activation rates were compared between groups for each neuron using a Mann-Whitney $U$ test. Rank-biserial correlation (RBC) quantified effect size. $P$-values were corrected using false discovery rate (FDR) \citep{benjamini_controlling_1995}, and reported as $q$ values, across all neurons and layers. Neurons with $q<0.05$ and higher activation rates for AD transcripts were pooled and ranked globally by AD-oriented RBC, yielding 67,640 eligible neurons. We defined three intervention scopes: the top 2,000, top 10,000, and all significant neurons, allowing intervention scope to vary while preserving the same ranking criterion. Their layer-wise distributions are reported in Appendix~\ref{sec:neuron_distribute}.

\subsection{Modulating AD-associated neurons}
We used Qwen3-8B, an 8-billion-parameter dense language model with 36 layers and 12,288 FFN units per layer \citep{yang_qwen3_2025}. Neuron modulation was implemented by scaling the down-projection weights associated with selected neurons. 

In the gated FFN, the gate activation vector $\mathbf{a}^{l}_{t}$ is combined element-wise with the up-projection output:

\begin{equation}
\mathbf{m}^{l}_{t}=
\mathbf{a}^{l}_{t}
\odot
\left(
\mathbf{W}^{l}_{\mathrm{up}}
\mathbf{h}^{l}_{t}
\right),
\end{equation}

where $\mathbf{W}^{l}_{\mathrm{up}}$ is the up-projection matrix and $\odot$ denotes element-wise multiplication. The scalar $m^{l}_{t,i}$ represents the effective output of neuron $i$ for token $t$ after the gate and up-projection branches have been combined.

The gated neuron outputs are projected back to the model hidden dimension through the down-projection matrix

\begin{equation}
\mathbf{o}^{l}_{t} = \mathbf{W}^{l}_{\mathrm{down}} \, \mathbf{m}^{l}_{t},
\end{equation}

where $\mathbf{W}^{l}_{\mathrm{down}}$ maps the FFN intermediate representation back to the model hidden dimension. Let $\mathbf{w}^{l}_{\mathrm{down},i}$ denote the output-weight vector associated with neuron $i$. The contribution of neuron $i$ to the FFN output is

\begin{equation}
\mathbf{c}^{l}_{t,i} = m^{l}_{t,i}\,\mathbf{w}^{l}_{\mathrm{down},i},
\end{equation}

For each neuron in the selected intervention set, we scaled its output-weight vector as

\begin{equation}
\mathbf{w}^{\prime l}_{\mathrm{down},i} = \exp(\alpha)\, \mathbf{w}^{l}_{\mathrm{down},i},
\end{equation}

where $\alpha$ controls the direction and magnitude of the intervention. Output-weight vectors associated with non-selected neurons were left unchanged. The contribution of a selected neuron after scaling is therefore

\begin{equation}
\mathbf{c}^{\prime l}_{t,i} = \exp(\alpha) \, \mathbf{c}^{l}_{t,i}.
\end{equation}

This procedure altered the effective output contributions of selected neurons without overwriting their gate activations.

We used $\alpha=0.2$ and $0.6$ for approximately $1.22$- and $1.82$-fold amplification, and $\alpha=-0.6$ for attenuation to approximately $0.55$ of the original contribution. Applying each magnitude to the top 2,000, top 10,000, and all significant neurons produced nine edited variants. Each was saved as a separate checkpoint derived from the unedited model.  Only the down-projection vectors associated with selected neurons were modified. All remaining parameters were identical to those of the original model.

\section{Experiment}
\label{sec:experiment}

\subsection{Experiment setup}
We evaluated the original Qwen3-8B model and nine edited variants, representing all combinations of three intervention magnitudes ($\alpha=-0.6$, $0.2$, and $0.6$) and three scopes (the top 2,000, the top 10,000, and all significant AD-associated neurons). Following an occupational role-playing approach to diversify LLM responses \citep{wang_fluency-based_2025}, we created 25 fictional retired-participant profiles and used them across all ten conditions, yielding 250 sessions (Appendix~\ref{sec:prompts}). Each session followed the same 12-turn sequence, with conversation history retained to support delayed recall. We applied the Qwen chat template, disabled explicit thinking, and allowed up to 1,024 new tokens per turn. All other parameters followed the checkpoint defaults.

\subsection{Neuropsychological battery and assessment}
The battery adapted established neuropsychological paradigms for characterizing cognitive dysfunction in AD, probing verbal episodic memory, lexical retrieval, working memory, procedural organization, episodic simulation, and referential processing. Full prompts are provided in Appendix~\ref{sec:prompts}. Two raters, blinded to model condition, independently scored the responses. Agreement was assessed using mean absolute error (MAE) and average-measures intraclass correlation (ICC2k). Analyses used the mean rating. The scoring scheme is provided in Appendix~\ref{sec:rating_scheme}.

\paragraph{Story recall.} The model first received a short narrative about a thirsty bird using pebbles to reach water, as available from the DISCOURSE protocol \citep{cho_discourse_2026}. It then recalled it immediately and after nine intervening tasks. Immediate and delayed story recall are established verbal paradigms to characterize episodic memory impairment in AD \citep{greene_analysis_1996}. Raters scored the presence of seven predefined entities and twelve events. The primary outcomes were the numbers of recalled entities and events.

\paragraph{Verbal fluency.} Immediate story recall was followed by two verbal-fluency tasks. Semantic and phonemic fluency are paradigms commonly used to characterize lexical semantics and executive dysfunction in AD \citep{tessaro_verbal_2020}, which required up to 15 animal names and 15 words beginning with C, respectively. We removed non-task content, collapsed morphological variants of the same item, and counted each valid response once. Animal-name validity was checked manually for semantic fluency, whereas C-initial word validity was checked automatically for phonemic fluency. The primary outcomes were the number of correct responses.

\paragraph{Working memory.} Working memory was assessed using digit-span-like tasks and mixed digit-letter-span-like tasks. Span tasks have previously been jointly used to examine verbal working-memory impairment in MCI and AD \citep{kessels_assessment_2011}. A response was correct only when complete sequence appeared in the required order. Errors were summed across four tasks, yielding a score from 0 to 4.

\paragraph{Procedural discourse.} Procedural discourse assessed access to script-based action knowledge and the ability to organize the component steps of a familiar everyday activity. Such discourse has previously been used to characterize the omission, repetition, and ordering of procedural information in AD \citep{ramsden_performance_2008}. In our battery, the model was asked to describe how to make a cup of tea. Raters scored the completeness of six essential steps and the overall coherence of ordering.

\paragraph{Scene construction.} Scene construction assessed constructive episodic simulation \citep{irish_scene_2015}. The model constructed a novel supermarket scene, rather than recalling an autobiographical event. Following \citet{he_episodic_2024}, raters classified each sentence or embedded clause as episodic (EP), non-episodic (NONEP), or OTHER. OTHER comprises mainly metalinguistic or non-task content such as sighs or "I don't know". Outcomes were the proportions of annotated units assigned to the three categories.

\paragraph{Coreference resolution.} Coreference resolution probes grammatical, lexical-semantic, and discourse integration, which is impaired across prodromal and clinical AD \citep{drummond_deficits_2015, lust_disintegration_2024}. The model was instructed to rewrite a short narrative by replacing each pronoun with its intended antecedent while preserving the original content and event order. Each of eight target pronouns received 1 for a correct antecedent, 0 if unresolved, and $-1$ if incorrect. The eight item scores were summed.

\subsection{Computational linguistic profiling}
In addition to human scoring, we analyzed immediate recall, delayed recall, and scene construction, with six computational linguistic measures, because they produced sufficiently extended, unconstrained discourse. These measures captured discourse quantity and lexical, informational, and syntactic organization. Token count indexed discourse quantity. Moving-average type-token ratio measured lexical richness. Frequency-based lexical surprisal indexed reliance on more or less frequent vocabulary. Lexical density and DEPID-R quantified complementary aspects of idea density. Lexical density measured the proportion of content words, whereas DEPID-R estimated the amount of non-redundant propositional dependency relations per token \citep{sirts_idea_2017}. Hierarchical syntactic complexity was assessed using the ratio of embedded clauses, which indexed the extent of clausal elaboration \citep{ivanova_defying_2023}. These measures target changes commonly reported in AD speech, including reduced output, lexical richness, idea density, and syntactic elaboration, and greater reliance on frequent words \citep{boschi_connected_2017, sirts_idea_2017, kave_word_2018, slegers_connected_2018, ivanova_defying_2023}. Definitions and preprocessing details are provided in Appendix~\ref{sec:nlp_feature}.

\subsection{Statistical analysis}
Each edited condition was compared with the unedited model using generalized estimating equations. Model condition was categorical, with the unedited model as reference. Repeated observations were clustered by occupational profile under an exchangeable correlation structure. Occupational profile was also included as a covariate, and response length was included for all automated linguistic outcomes except token count. Families and link functions followed outcome scale. The nine comparisons within each outcome were FDR-corrected.

\begin{figure*}
  \centering
  \includegraphics[width=0.95\linewidth]{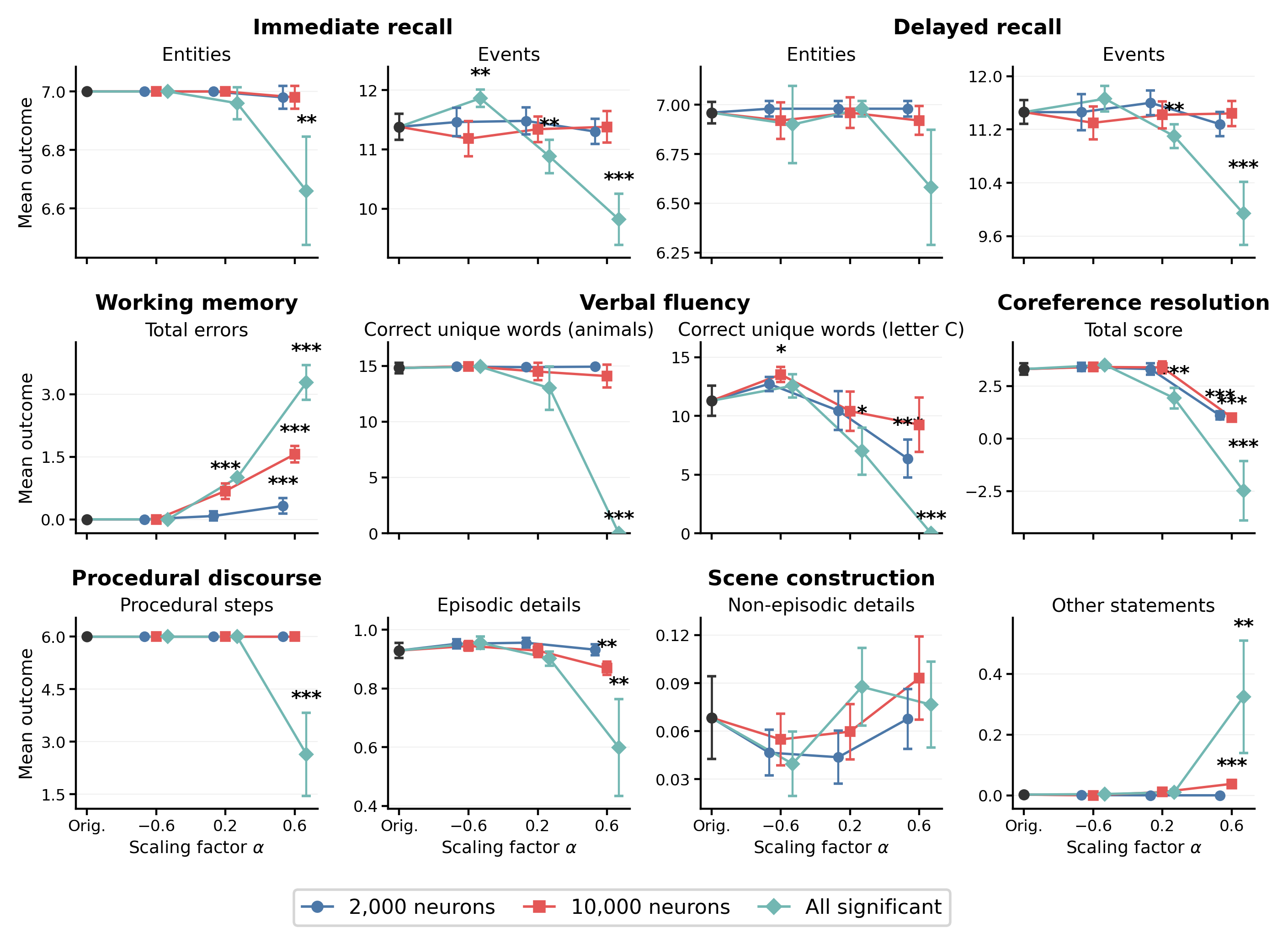} 
  \caption{\textbf{Effects on task perfromance.}
  Points show condition means with 95\% confidence intervals. Black point denotes the original Qwen3-8B model. Edited conditions vary by intervention scope and scaling factor. Asterisks indicate comparisons with the original model: $^{*}q<0.05$, $^{**}q<0.01$, $^{***}q<0.001$.}
  \label{fig:core_outcomes}
\end{figure*}

\begin{figure*}
  \centering
  \includegraphics[width=\linewidth]{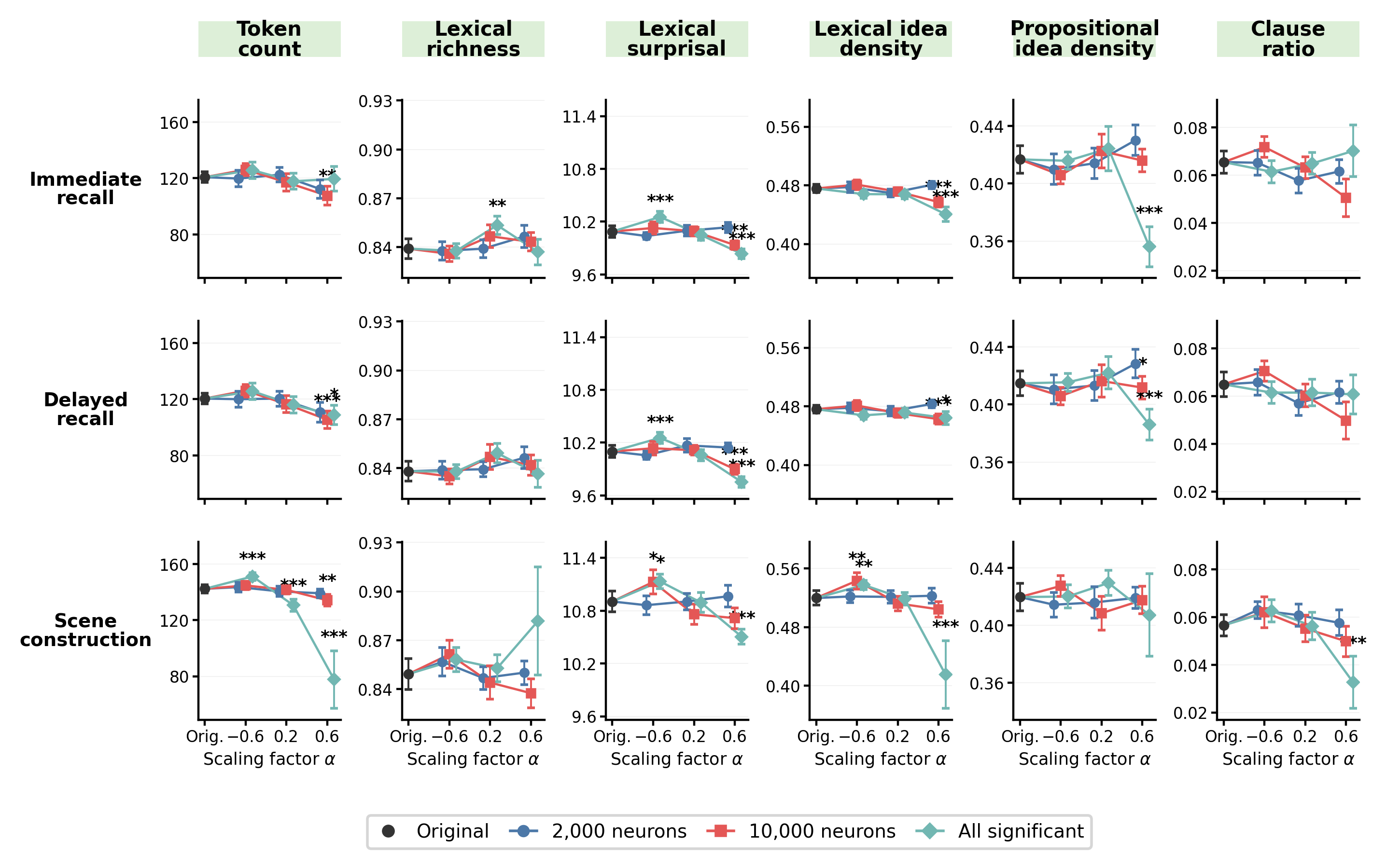} 
  \caption{\textbf{Effects on linguistic measures.}
  Rows show immediate recall, delayed recall, and scene construction. Columns show six linguistic measures. Points and error bars represent condition means and 95\% confidence intervals. Asterisks indicate comparisons with the original model: $^{*}q<0.05$, $^{**}q<0.01$,  $^{***}q<0.001$.}
  \label{fig:linguistic_outcomes}
\end{figure*}

\section{Results}
\paragraph{Inter-rater reliability for human ratings.} Agreement was moderate to excellent for event count in  story recall, procedural discourse, scene construction, and coreference resolution (Table~\ref{tab:annotation_agreement}). ICCs for entity count in story recall were near zero because responses showed little between-session variability, with most reaching the maximum score of 7. The corresponding MAEs were nevertheless very small, indicating that disagreements were rare and minor.

\begin{table}[tbhp]
\centering
\setlength{\tabcolsep}{2.5pt}
\begin{tabular}{llrr}
\hline
Task & Measure & ICC2k & MAE \\
\hline
Immediate recall & Entities\# & -0.009 & 0.084 \\
                 & Events\# & 0.825 & 0.360 \\
Delayed recall & Entities\# & 0.029 & 0.160 \\
               & Events\# & 0.630 & 0.364 \\
Procedural discourse & Step & 1.000 & 0.000 \\
                     & Order & 1.000 & 0.000 \\
Scene construction & EP\% & 0.967 & 0.046 \\
                   & NONEP\% & 0.686 & 0.046 \\
                   & OTHER\% & 1.000 & 0.002 \\
Coreference resolution & Score & 0.971 & 0.588 \\
\hline
\end{tabular}
\caption{Inter-rater agreement for manual annotations.}
\label{tab:annotation_agreement}
\end{table}

\paragraph{Neuron modulation produces direction- and scope-dependent changes.}
The strongest amplification produced the largest and broadest performance changes, but significant effects were not confined to this condition. Across conditions, neuron modulation showed direction-, magnitude-, and scope-dependent effects across all assessed domains, broadly paralleling impairments reported in human AD (Figure~\ref{fig:core_outcomes}). Attenuation ($\alpha=-0.6$) left most outcomes unchanged and selectively improved immediate event recall and phonemic fluency. Moderate amplification ($\alpha=0.2$) already affected story recall, phonemic fluency, working memory, and coreference resolution. The broadest impairments occurred when all significant neurons were amplified at $\alpha=0.6$, affecting immediate and delayed recall, semantic and phonemic fluency, working memory, procedural discourse, scene construction, and coreference resolution. The sequence-order score in procedural discourse followed the same pattern as procedural-step completeness and is omitted from the figure to avoid redundancy. In scene construction, the reduction in the episodic-detail ratio was accompanied by an increase in OTHER statements, whereas the non-episodic-detail ratio did not change significantly, mirroring the redistribution reported previously \citep{he_episodic_2024}. Smaller intervention sets produced narrower effects, primarily on phonemic fluency, working memory, the episodic-detail ratio, and coreference resolution. The effects therefore reflected a structured response to modulation rather than nonspecific model degradation.

\paragraph{Linguistic consequences of neuron modulation.} Linguistic effects were similarly task-, magnitude-, and scope-dependent (Figure~\ref{fig:linguistic_outcomes}). Strong amplification of all AD-associated neurons ($\alpha=0.6$) reduced lexical surprisal and both idea density measures in immediate recall. Delayed recall showed lower lexical surprisal, propositional idea density, and response length, whereas scene construction showed the broadest reductions in response length, lexical surprisal, lexical idea density, and clause ratio. Lexical richness was largely preserved, apart from an increase in immediate recall. Attenuation selectively increased word count, lexical surprisal, and lexical idea density during scene construction, while moderate amplification produced smaller and less consistent effects. The selective profile suggests reorganization of generated language rather than uniform degradation in quality.

\section{Discussion}
We introduced an activation-guided framework for identifying and editing LLM neurons associated with AD language. Amplification produced progressively broader impairments as magnitude and scope increased, whereas attenuation largely preserved performance and selectively improved some outcomes. The strongest intervention affected recall, fluency, working memory, procedural discourse, scene construction, and coreference, alongside lower word count, lexical surprisal, idea density, and syntactic complexity. Together, these findings establish a reproducible link between AD-associated internal representations and downstream model behavior.

Our aim was not only to reproduce AD-like behavior, but to construct an experimentally manipulable model for a behavioral phenotype. Although both intervention targets and LLM pretraining were language-grounded, modulation affected tasks conventionally assigned to non-language domains, such as episodic memory, working memory, and executive function \citep{shallice_neuropsychology_1988}. These findings challenge accounts that assign language primarily a communicative function \citep{fedorenko_language_2024} and, more broadly, the traditional partition of cognition into separable domains \citep{shallice_neuropsychology_1988}. Within the model, language-sensitive representations contributed to maintaining, retrieving, and organizing information across these conventional boundaries, rather than functioning only as an output channel. This interpretation remains constrained by the fact that all tasks were administered and answered through language.

The directional and graded effects further argue against indiscriminate model degradation. Amplification progressively expanded the impairment profile, whereas attenuation did not produce a corresponding global disruption. The intervention instead appears to shift a distributed computational system along a structured behavioral dimension, with different capacities becoming vulnerable at different levels of perturbation. The accompanying linguistic changes likewise suggest a selective reorganization of generated language rather than a uniform decline in output quality. Post hoc, layer-matched random interventions provided further evidence that observed effects were not attributable to nonspecific perturbation alone (Appendix~\ref{sec:random_control}).

In this limited sense, the edited model offers an experimental role loosely analogous to the function of an animal model. Language disorders lack an animal model capable of reproducing the relevant linguistic phenotypes, while many causal manipulations are infeasible or unethical in patients. Qwen3-8B does not have AD, and its artificial neurons are not biological neurons. Yet, its value lies in providing a tractable test bed in which differences derived from patient language can be localized, selectively amplified or attenuated, and linked to downstream behavioral consequences under controlled conditions. Such models may therefore serve as pilot systems for identifying candidate mechanisms and intervention hypotheses before being examined in human. The relevant resemblance is not between the model and the disease as wholes, but their experimentally testable structures of dependence.

\section{Conclusion}
We presented an activation-guided neuron intervention framework for testing whether neurons associated with AD language can causally influence LLM behavior. Selectively modulating these neurons produced systematic changes across neuropsychological battery, with amplification shifting Qwen3-8B toward an AD-related computational language phenotype. These results support neuron-level intervention as a controlled approach for studying clinically relevant language behavior in LLMs, without implying that the model itself has AD.

\section*{Limitations}
This study used a single language model and one English clinical dataset, and all tasks were administered and answered through language. These design choices limit cross-model and cross-linguistic generalizability. Future work should test whether the observed effects replicate across model architectures, languages, clinical cohorts, other diagnoses, and task formats, and whether they track patient-level variation in disease stage, symptom severity, and biological markers.

\section*{Ethical Considerations}
The ADReSSo data used in this study were obtained through DementiaBank under its controlled-access agreement. The original DementiaBank data were collected with informed consent from participants and approved by the relevant ethics committee, with details provided in \citep{luz_detecting_2021}. The present study involved secondary analysis of de-identified, controlled-access transcripts and did not involve new data collection or participant contact. We fully complied with all DementiaBank requirements regarding data access, storage, use, and confidentiality. The transcripts were used solely for the approved research purposes and were not redistributed or made publicly available. Any released code or derived resources will exclude the original DementiaBank data, and users must obtain independent access through the official DementiaBank application process. This work is intended as a controlled computational investigation rather than a clinical model or diagnostic tool. A feed-forward unit in a language model is not a biological neuron, and the induced outputs should not be interpreted as evidence that the model has Alzheimer’s disease or reproduces the disorder itself. Overinterpretation of these results could promote reductive or stereotypical representations of people with Alzheimer’s disease, and the intervention method could potentially be misused to generate stereotyped disease-related language. We therefore restrict our claims to computational language phenotypes observed under controlled model interventions and do not advocate clinical deployment, diagnosis, or individual-level inference.

\section*{Acknowledgments}
We thank the raters for their contributions in response evaluation. This study was funded by European Research Council (ERC-2023-SyG, 101118756) and Department of Science and Technology of Guangdong Province (2023A0505050118). The views and opinions expressed are those of the authors only and do not necessarily reflect those of the European Union.


\bibliography{custom}

@misc{yang_qwen3_2025,
    title = {Qwen3 {Technical} {Report}},
    url = {http://arxiv.org/abs/2505.09388},
    doi = {10.48550/arXiv.2505.09388},
    urldate = {2026-04-22},
    publisher = {arXiv},
    author = {Yang, An and Li, Anfeng and Yang, Baosong and Zhang, Beichen and Hui, Binyuan and Zheng, Bo and Yu, Bowen and Gao, Chang and Huang, Chengen and Lv, Chenxu and Zheng, Chujie and Liu, Dayiheng and Zhou, Fan and Huang, Fei and Hu, Feng and Ge, Hao and Wei, Haoran and Lin, Huan and Tang, Jialong and Yang, Jian and Tu, Jianhong and Zhang, Jianwei and Yang, Jianxin and Yang, Jiaxi and Zhou, Jing and Zhou, Jingren and Lin, Junyang and Dang, Kai and Bao, Keqin and Yang, Kexin and Yu, Le and Deng, Lianghao and Li, Mei and Xue, Mingfeng and Li, Mingze and Zhang, Pei and Wang, Peng and Zhu, Qin and Men, Rui and Gao, Ruize and Liu, Shixuan and Luo, Shuang and Li, Tianhao and Tang, Tianyi and Yin, Wenbiao and Ren, Xingzhang and Wang, Xinyu and Zhang, Xinyu and Ren, Xuancheng and Fan, Yang and Su, Yang and Zhang, Yichang and Zhang, Yinger and Wan, Yu and Liu, Yuqiong and Wang, Zekun and Cui, Zeyu and Zhang, Zhenru and Zhou, Zhipeng and Qiu, Zihan},
    month = may,
    year = {2025},
    note = {arXiv:2505.09388 [cs]},
}

@article{shankar_systematic_2025,
    title = {A {Systematic} {Review} of {Natural} {Language} {Processing} {Techniques} for {Early} {Detection} of {Cognitive} {Impairment}},
    volume = {3},
    issn = {2949-7612},
    url = {https://www.sciencedirect.com/science/article/pii/S2949761225000124},
    doi = {10.1016/j.mcpdig.2025.100205},
    number = {2},
    urldate = {2026-07-24},
    journal = {Mayo Clinic Proceedings: Digital Health},
    author = {Shankar, Ravi and Bundele, Anjali and Mukhopadhyay, Amartya},
    month = jun,
    year = {2025},
    pages = {100205},
}

@article{van_den_berg_digital_2024,
    title = {Digital remote assessment of speech acoustics in cognitively unimpaired adults: feasibility, reliability and associations with amyloid pathology},
    volume = {16},
    issn = {1758-9193},
    shorttitle = {Digital remote assessment of speech acoustics in cognitively unimpaired adults},
    url = {https://doi.org/10.1186/s13195-024-01543-3},
    doi = {10.1186/s13195-024-01543-3},
    language = {en},
    number = {1},
    urldate = {2026-07-24},
    journal = {Alzheimer's Research \& Therapy},
    author = {van den Berg, Rosanne L. and de Boer, Casper and Zwan, Marissa D. and Jutten, Roos J. and van Liere, Mariska and van de Glind, Marie-Christine A.B.J. and Dubbelman, Mark A. and Schlüter, Lisa Marie and van Harten, Argonde C. and Teunissen, Charlotte E. and van de Giessen, Elsmarieke and Barkhof, Frederik and Collij, Lyduine E. and Robin, Jessica and Simpson, William and Harrison, John E. and van der Flier, Wiesje M. and Sikkes, Sietske A.M.},
    month = aug,
    year = {2024},
    pages = {176},
}

@article{he_automated_2023,
    title = {Automated {Classification} of {Cognitive} {Decline} and {Probable} {Alzheimer}'s {Dementia} {Across} {Multiple} {Speech} and {Language} {Domains}},
    volume = {32},
    issn = {1058-0360, 1558-9110},
    url = {http://pubs.asha.org/doi/10.1044/2023_AJSLP-22-00403},
    doi = {10.1044/2023_AJSLP-22-00403},
    language = {en},
    number = {5},
    urldate = {2026-07-24},
    journal = {American Journal of Speech-Language Pathology},
    author = {He, Rui and Chapin, Kayla and Al-Tamimi, Jalal and Bel, Núria and Marquié, Marta and Rosende-Roca, Maitee and Pytel, Vanesa and Tartari, Juan Pablo and Alegret, Montse and Sanabria, Angela and Ruiz, Agustín and Boada, Mercè and Valero, Sergi and Hinzen, Wolfram},
    month = sep,
    year = {2023},
    pages = {2075--2086},
}

@article{fristed_leveraging_2022,
    title = {Leveraging speech and artificial intelligence to screen for early {Alzheimer}’s disease and amyloid beta positivity},
    volume = {4},
    issn = {2632-1297},
    url = {https://doi.org/10.1093/braincomms/fcac231},
    doi = {10.1093/braincomms/fcac231},
    number = {5},
    urldate = {2026-07-24},
    journal = {Brain Communications},
    author = {Fristed, Emil and Skirrow, Caroline and Meszaros, Marton and Lenain, Raphael and Meepegama, Udeepa and Papp, Kathryn V and Ropacki, Michael and Weston, Jack},
    month = oct,
    year = {2022},
    pages = {fcac231},
}

@article{robin_evaluation_2020,
    title = {Evaluation of {Speech}-{Based} {Digital} {Biomarkers}: {Review} and {Recommendations}},
    volume = {4},
    issn = {2504-110X},
    shorttitle = {Evaluation of {Speech}-{Based} {Digital} {Biomarkers}},
    url = {https://pmc.ncbi.nlm.nih.gov/articles/PMC7670321/},
    doi = {10.1159/000510820},
    number = {3},
    urldate = {2026-07-24},
    journal = {Digital Biomarkers},
    author = {Robin, Jessica and Harrison, John E. and Kaufman, Liam D. and Rudzicz, Frank and Simpson, William and Yancheva, Maria},
    month = oct,
    year = {2020},
    pages = {99--108},
}

@article{fraser_linguistic_2015,
    title = {Linguistic {Features} {Identify} {Alzheimer}’s {Disease} in {Narrative} {Speech}},
    volume = {49},
    issn = {13872877, 18758908},
    url = {https://www.medra.org/servlet/aliasResolver?alias=iospress&doi=10.3233/JAD-150520},
    doi = {10.3233/JAD-150520},
    language = {en},
    number = {2},
    urldate = {2021-03-14},
    journal = {Journal of Alzheimer's Disease},
    author = {Fraser, Kathleen C. and Meltzer, Jed A. and Rudzicz, Frank},
    editor = {Garrard, Peter},
    month = oct,
    year = {2015},
    pages = {407--422},
}

@inproceedings{luz_alzheimers_2020,
    title = {Alzheimer’s {Dementia} {Recognition} {Through} {Spontaneous} {Speech}: {The} {ADReSS} {Challenge}},
    shorttitle = {Alzheimer’s {Dementia} {Recognition} {Through} {Spontaneous} {Speech}},
    url = {http://www.isca-speech.org/archive/Interspeech_2020/abstracts/2571.html},
    doi = {10.21437/Interspeech.2020-2571},
    language = {en},
    urldate = {2021-06-29},
    booktitle = {Interspeech 2020},
    publisher = {ISCA},
    author = {Luz, Saturnino and Haider, Fasih and Fuente, Sofia de la and Fromm, Davida and MacWhinney, Brian},
    month = oct,
    year = {2020},
    pages = {2172--2176},
}

@inproceedings{balagopalan_bert_2020,
    title = {To {BERT} or not to {BERT}: {Comparing} {Speech} and {Language}-{Based} {Approaches} for {Alzheimer}’s {Disease} {Detection}},
    shorttitle = {To {BERT} or not to {BERT}},
    url = {http://www.isca-speech.org/archive/Interspeech_2020/abstracts/2557.html},
    doi = {10.21437/Interspeech.2020-2557},
    language = {en},
    urldate = {2021-06-29},
    booktitle = {Interspeech 2020},
    publisher = {ISCA},
    author = {Balagopalan, Aparna and Eyre, Benjamin and Rudzicz, Frank and Novikova, Jekaterina},
    month = oct,
    year = {2020},
    pages = {2167--2171},
}

@inproceedings{farzana_domain_2024,
    address = {Miami, Florida, USA},
    title = {Domain {Adaptation} via {Prompt} {Learning} for {Alzheimer}'s {Detection}},
    url = {https://aclanthology.org/2024.findings-emnlp.937/},
    doi = {10.18653/v1/2024.findings-emnlp.937},
    urldate = {2026-07-24},
    booktitle = {Findings of the {Association} for {Computational} {Linguistics}: {EMNLP} 2024},
    publisher = {Association for Computational Linguistics},
    author = {Farzana, Shahla and Parde, Natalie},
    editor = {Al-Onaizan, Yaser and Bansal, Mohit and Chen, Yun-Nung},
    month = nov,
    year = {2024},
    pages = {15963--15976},
}

@inproceedings{heitz_linguistic_2025,
    address = {Abu Dhabi, UAE},
    title = {Linguistic {Features} {Extracted} by {GPT}-4 {Improve} {Alzheimer}'s {Disease} {Detection} based on {Spontaneous} {Speech}},
    url = {https://aclanthology.org/2025.coling-main.126/},
    urldate = {2026-07-24},
    booktitle = {Proceedings of the 31st {International} {Conference} on {Computational} {Linguistics}},
    publisher = {Association for Computational Linguistics},
    author = {Heitz, Jonathan and Schneider, Gerold and Langer, Nicolas},
    editor = {Rambow, Owen and Wanner, Leo and Apidianaki, Marianna and Al-Khalifa, Hend and Eugenio, Barbara Di and Schockaert, Steven},
    month = jan,
    year = {2025},
    pages = {1850--1864},
}

@article{colla_semantic_2022,
    title = {Semantic coherence markers: {The} contribution of perplexity metrics},
    volume = {134},
    issn = {0933-3657},
    shorttitle = {Semantic coherence markers},
    url = {https://www.sciencedirect.com/science/article/pii/S0933365722001440},
    doi = {10.1016/j.artmed.2022.102393},
    language = {en},
    urldate = {2023-05-10},
    journal = {Artificial Intelligence in Medicine},
    author = {Colla, Davide and Delsanto, Matteo and Agosto, Marco and Vitiello, Benedetto and Radicioni, Daniele P.},
    month = dec,
    year = {2022},
    pages = {102393},
}

@article{jiang_structure_2025,
    title = {The structure of spontaneous speech changes in {Alzheimer}’s disease: {Crosslingual} evidence from {English} and {Greek}},
    volume = {20},
    issn = {1932-6203},
    shorttitle = {The structure of spontaneous speech changes in {Alzheimer}’s disease},
    url = {https://journals.plos.org/plosone/article?id=10.1371/journal.pone.0324270},
    doi = {10.1371/journal.pone.0324270},
    language = {en},
    number = {5},
    urldate = {2025-07-20},
    journal = {PLOS ONE},
    publisher = {Public Library of Science},
    author = {Jiang, Hong and Chen, Zhengwei and Liu, Yu and Yang, Chun and Yuan, Xiaofeng and He, Rui},
    month = may,
    year = {2025},
    pages = {e0324270},
}

@inproceedings{gkoumas-etal-2023-digital,
    title = "A Digital Language Coherence Marker for Monitoring Dementia",
    author = "Gkoumas, Dimitris  and
      Tsakalidis, Adam  and
      Liakata, Maria",
    editor = "Bouamor, Houda  and
      Pino, Juan  and
      Bali, Kalika",
    booktitle = "Proceedings of the 2023 Conference on Empirical Methods in Natural Language Processing",
    month = dec,
    year = "2023",
    address = "Singapore",
    publisher = "Association for Computational Linguistics",
    url = "https://aclanthology.org/2023.emnlp-main.994/",
    doi = "10.18653/v1/2023.emnlp-main.994",
    pages = "16021--16034"
}

@inproceedings{jawahar_what_2019,
    address = {Florence, Italy},
    title = {What {Does} {BERT} {Learn} about the {Structure} of {Language}?},
    url = {https://aclanthology.org/P19-1356},
    doi = {10.18653/v1/P19-1356},
    urldate = {2022-09-28},
    booktitle = {Proceedings of the 57th {Annual} {Meeting} of the {Association} for {Computational} {Linguistics}},
    publisher = {Association for Computational Linguistics},
    author = {Jawahar, Ganesh and Sagot, Benoît and Seddah, Djamé},
    year = {2019},
    pages = {3651--3657},
}

@inproceedings{xiao_neuron-based_2025,
    title = {Neuron-based {Multifractal} {Analysis} of {Neuron} {Interaction} {Dynamics} in {Large} {Models}},
    volume = {2025},
    url = {https://proceedings.iclr.cc/paper_files/paper/2025/file/6158e152498f8d8b83d14388a7ec1963-Paper-Conference.pdf},
    booktitle = {International {Conference} on {Learning} {Representations}},
    author = {Xiao, Xiongye and Ping, Heng and Zhou, Chenyu and Cao, Defu and Li, Yaxing and Zhou, Yi-Zhuo and Li, Shixuan and Kanakaris, Nikos and Bogdan, Paul},
    editor = {Yue, Y. and Garg, A. and Peng, N. and Sha, F. and Yu, R.},
    year = {2025},
    pages = {39025--39072},
}

@inproceedings{lai_style-specific_2024,
    address = {Miami, Florida, USA},
    title = {Style-{Specific} {Neurons} for {Steering} {LLMs} in {Text} {Style} {Transfer}},
    url = {https://aclanthology.org/2024.emnlp-main.745/},
    doi = {10.18653/v1/2024.emnlp-main.745},
    urldate = {2026-07-24},
    booktitle = {Proceedings of the 2024 {Conference} on {Empirical} {Methods} in {Natural} {Language} {Processing}},
    publisher = {Association for Computational Linguistics},
    author = {Lai, Wen and Hangya, Viktor and Fraser, Alexander},
    editor = {Al-Onaizan, Yaser and Bansal, Mohit and Chen, Yun-Nung},
    month = nov,
    year = {2024},
    pages = {13427--13443},
}

@inproceedings{dai-etal-2022-knowledge,
    title = "Knowledge Neurons in Pretrained Transformers",
    author = "Dai, Damai  and
      Dong, Li  and
      Hao, Yaru  and
      Sui, Zhifang  and
      Chang, Baobao  and
      Wei, Furu",
    editor = "Muresan, Smaranda  and
      Nakov, Preslav  and
      Villavicencio, Aline",
    booktitle = "Proceedings of the 60th Annual Meeting of the Association for Computational Linguistics (Volume 1: Long Papers)",
    month = may,
    year = "2022",
    address = "Dublin, Ireland",
    publisher = "Association for Computational Linguistics",
    url = "https://aclanthology.org/2022.acl-long.581/",
    doi = "10.18653/v1/2022.acl-long.581",
    pages = "8493--8502"
}

@inproceedings{mueller-etal-2022-causal,
    title = "Causal Analysis of Syntactic Agreement Neurons in Multilingual Language Models",
    author = "Mueller, Aaron  and
      Xia, Yu  and
      Linzen, Tal",
    editor = "Fokkens, Antske  and
      Srikumar, Vivek",
    booktitle = "Proceedings of the 26th Conference on Computational Natural Language Learning (CoNLL)",
    month = dec,
    year = "2022",
    address = "Abu Dhabi, United Arab Emirates (Hybrid)",
    publisher = "Association for Computational Linguistics",
    url = "https://aclanthology.org/2022.conll-1.8/",
    doi = "10.18653/v1/2022.conll-1.8",
    pages = "95--109"
}

@inproceedings{li_gpt-d_2022,
    address = {Dublin, Ireland},
    title = {{GPT}-{D}: {Inducing} {Dementia}-related {Linguistic} {Anomalies} by {Deliberate} {Degradation} of {Artificial} {Neural} {Language} {Models}},
    shorttitle = {{GPT}-{D}},
    url = {https://aclanthology.org/2022.acl-long.131/},
    doi = {10.18653/v1/2022.acl-long.131},
    urldate = {2026-07-24},
    booktitle = {Proceedings of the 60th {Annual} {Meeting} of the {Association} for {Computational} {Linguistics} ({Volume} 1: {Long} {Papers})},
    publisher = {Association for Computational Linguistics},
    author = {Li, Changye and Knopman, David and Xu, Weizhe and Cohen, Trevor and Pakhomov, Serguei},
    editor = {Muresan, Smaranda and Nakov, Preslav and Villavicencio, Aline},
    month = may,
    year = {2022},
    pages = {1866--1877},
}

@inproceedings{li_too_2024,
    address = {Bangkok, Thailand},
    title = {Too {Big} to {Fail}: {Larger} {Language} {Models} are {Disproportionately} {Resilient} to {Induction} of {Dementia}-{Related} {Linguistic} {Anomalies}},
    shorttitle = {Too {Big} to {Fail}},
    url = {https://aclanthology.org/2024.findings-acl.380/},
    doi = {10.18653/v1/2024.findings-acl.380},
    urldate = {2026-07-24},
    booktitle = {Findings of the {Association} for {Computational} {Linguistics}: {ACL} 2024},
    publisher = {Association for Computational Linguistics},
    author = {Li, Changye and Sheng, Zhecheng and Cohen, Trevor and Pakhomov, Serguei},
    editor = {Ku, Lun-Wei and Martins, Andre and Srikumar, Vivek},
    month = aug,
    year = {2024},
    pages = {6363--6377},
}

@misc{yang_lesioned_2026,
    title = {Lesioned {Multimodal} {Language} {Models} {Reproduce} {Aphasic} {Picture}-{Naming} {Patterns}},
    url = {http://arxiv.org/abs/2607.11621},
    doi = {10.48550/arXiv.2607.11621},
    urldate = {2026-07-24},
    publisher = {arXiv},
    author = {Yang, Yong and Guan, Xiang and Arheix-Parras, Sophie and Ahmadi, Saeed and Newman-Norlund, Roger and Bonilha, Leonardo and Rorden, Christopher and Fridriksson, Julius and Desai, Rutvik H. and Nelakuditi, Srihari},
    month = jul,
    year = {2026},
    note = {arXiv:2607.11621 [cs.AI]},
}

@misc{roll_artificial_2026,
    title = {Artificial {Aphasias} in {Lesioned} {Language} {Models}},
    url = {http://arxiv.org/abs/2605.16222},
    doi = {10.48550/arXiv.2605.16222},
    urldate = {2026-07-24},
    publisher = {arXiv},
    author = {Roll, Nathan and Kries, Jill and Gwilliams, Laura and Shain, Cory},
    month = may,
    year = {2026},
    note = {arXiv:2605.16222 [cs.CL]},
}

@misc{wang_component-level_2026,
    title = {Component-{Level} {Lesioning} of {Language} {Models} {Reveals} {Clinically} {Aligned} {Aphasia} {Phenotypes}},
    url = {http://arxiv.org/abs/2601.19723},
    doi = {10.48550/arXiv.2601.19723},
    urldate = {2026-07-24},
    publisher = {arXiv},
    author = {Wang, Yifan and Zheng, Jichen and Sun, Jingyuan and Zhang, Yunhao and Ye, Chunyu and Li, Jixing and Zong, Chengqing and Wang, Shaonan},
    month = jan,
    year = {2026},
    note = {arXiv:2601.19723 [cs.CL]},
}

@misc{bau_identifying_2018,
    title = {Identifying and {Controlling} {Important} {Neurons} in {Neural} {Machine} {Translation}},
    url = {http://arxiv.org/abs/1811.01157},
    doi = {10.48550/arXiv.1811.01157},
    urldate = {2026-07-24},
    publisher = {arXiv},
    author = {Bau, Anthony and Belinkov, Yonatan and Sajjad, Hassan and Durrani, Nadir and Dalvi, Fahim and Glass, James},
    month = nov,
    year = {2018},
    note = {arXiv:1811.01157 [cs.CL]},
}

@article{benjamini_controlling_1995,
    title = {Controlling the {False} {Discovery} {Rate}: {A} {Practical} and {Powerful} {Approach} to {Multiple} {Testing}},
    volume = {57},
    issn = {2517-6161},
    shorttitle = {Controlling the {False} {Discovery} {Rate}},
    url = {https://onlinelibrary.wiley.com/doi/abs/10.1111/j.2517-6161.1995.tb02031.x},
    doi = {10.1111/j.2517-6161.1995.tb02031.x},
    language = {en},
    number = {1},
    urldate = {2023-01-17},
    journal = {Journal of the Royal Statistical Society: Series B (Methodological)},
    author = {Benjamini, Yoav and Hochberg, Yosef},
    year = {1995},
    pages = {289--300},
}

@article{wang_fluency-based_2025,
    title = {The fluency-based semantic network of {LLMs} differs from humans},
    volume = {3},
    issn = {2949-8821},
    url = {https://www.sciencedirect.com/science/article/pii/S294988212400063X},
    doi = {10.1016/j.chbah.2024.100103},
    urldate = {2026-07-24},
    journal = {Computers in Human Behavior: Artificial Humans},
    author = {Wang, Ye and Deng, Yaling and Wang, Ge and Li, Tong and Xiao, Hongjiang and Zhang, Yuan},
    month = mar,
    year = {2025},
    pages = {100103},
}

@misc{luz_detecting_2021,
    title = {Detecting cognitive decline using speech only: {The} {ADReSSo} {Challenge}},
    shorttitle = {Detecting cognitive decline using speech only},
    url = {http://arxiv.org/abs/2104.09356},
    doi = {10.48550/arXiv.2104.09356},
    urldate = {2022-09-27},
    publisher = {arXiv},
    author = {Luz, Saturnino and Haider, Fasih and de la Fuente, Sofia and Fromm, Davida and MacWhinney, Brian},
    month = mar,
    year = {2021},
    note = {arXiv:2104.09356 [cs, eess]},
}

@article{greene_analysis_1996,
    title = {Analysis of the episodic memory deficit in early {Alzheimer}'s disease: evidence from the doors and people test},
    volume = {34},
    issn = {0028-3932},
    shorttitle = {Analysis of the episodic memory deficit in early {Alzheimer}'s disease},
    doi = {10.1016/0028-3932(95)00151-4},
    language = {eng},
    number = {6},
    journal = {Neuropsychologia},
    author = {Greene, J. D. and Baddeley, A. D. and Hodges, J. R.},
    month = jun,
    year = {1996},
    pages = {537--551},
}

@article{cho_discourse_2026,
    title = {The {DISCOURSE} in psychosis ({London} {Ontario}): {A} speech dataset to examine communication disturbances in early-stage psychosis},
    volume = {65},
    issn = {2352-3409},
    shorttitle = {The {DISCOURSE} in psychosis ({London} {Ontario})},
    url = {https://pmc.ncbi.nlm.nih.gov/articles/PMC12907862/},
    doi = {10.1016/j.dib.2026.112517},
    urldate = {2026-05-04},
    journal = {Data in Brief},
    author = {Cho, Brian and Balles, Estée and Mackinley, Michael and Dzialoszynski, Paulina and Ford, Sabrina and Lodhi, Rohit and Palaniyappan, Lena},
    month = jan,
    year = {2026},
    pages = {112517},
}

@article{tessaro_verbal_2020,
    title = {Verbal fluency in {Alzheimer}'s disease and mild cognitive impairment in individuals with low educational level and its relationship with reading and writing habits},
    volume = {14},
    issn = {1980-5764},
    doi = {10.1590/1980-57642020dn14-030011},
    language = {eng},
    number = {3},
    journal = {Dementia \& Neuropsychologia},
    author = {Tessaro, Bruna and Hermes-Pereira, Andressa and Schilling, Lucas Porcello and Fonseca, Rochele Paz and Kochhann, Renata and Hübner, Lilian Cristine},
    year = {2020},
    pages = {300--307},
}

@article{kessels_assessment_2011,
    title = {Assessment of working-memory deficits in patients with mild cognitive impairment and {Alzheimer}'s dementia using {Wechsler}'s {Working} {Memory} {Index}},
    volume = {23},
    issn = {1594-0667},
    doi = {10.1007/BF03325245},
    language = {eng},
    number = {5-6},
    journal = {Aging Clinical and Experimental Research},
    author = {Kessels, Roy P. C. and Molleman, Pieter W. and Oosterman, Joukje M.},
    year = {2011},
    pages = {487--490},
}

@article{ramsden_performance_2008,
    title = {Performance of {Everyday} {Actions} in {Mild} {Alzheimer}'s {Disease}},
    volume = {22},
    doi = {10.1037/0894-4105.22.1.17},
    journal = {Neuropsychology},
    author = {Ramsden, Clare and Kinsella, Glynda and Ong, Ben and Storey, Elsdon},
    month = jan,
    year = {2008},
    pages = {17--26},
}

@article{irish_scene_2015,
    title = {Scene construction impairments in {Alzheimer}'s disease - {A} unique role for the posterior cingulate cortex},
    volume = {73},
    issn = {1973-8102},
    doi = {10.1016/j.cortex.2015.08.004},
    language = {eng},
    journal = {Cortex; a Journal Devoted to the Study of the Nervous System and Behavior},
    author = {Irish, Muireann and Halena, Stephanie and Kamminga, Jody and Tu, Sicong and Hornberger, Michael and Hodges, John R.},
    month = dec,
    year = {2015},
    pages = {10--23},
}

@article{he_episodic_2024,
    title = {Episodic {Thinking} in {Alzheimer}'s {Disease} {Through} the {Lens} of {Language}: {Linguistic} {Analysis} and {Transformer}-{Based} {Classification}},
    volume = {33},
    issn = {1558-9110},
    shorttitle = {Episodic {Thinking} in {Alzheimer}'s {Disease} {Through} the {Lens} of {Language}},
    doi = {10.1044/2023_AJSLP-23-00066},
    language = {eng},
    number = {1},
    journal = {American Journal of Speech-Language Pathology},
    author = {He, Rui and Yuan, Xiaofeng and Hinzen, Wolfram},
    month = jan,
    year = {2024},
    pages = {87--95},
}

@article{drummond_deficits_2015,
    title = {Deficits in narrative discourse elicited by visual stimuli are already present in patients with mild cognitive impairment},
    volume = {7},
    issn = {1663-4365},
    url = {https://www.frontiersin.org/journals/aging-neuroscience/articles/10.3389/fnagi.2015.00096/full},
    doi = {10.3389/fnagi.2015.00096},
    language = {English},
    urldate = {2026-07-25},
    journal = {Frontiers in Aging Neuroscience},
    publisher = {Frontiers},
    author = {Drummond, Cláudia and Coutinho, Gabriel and Fonseca, Rochele Paz and Assunção, Naima and Teldeschi, Alina and de Oliveira-Souza, Ricardo and Moll, Jorge and Tovar-Moll, Fernanda and Mattos, Paulo},
    month = may,
    year = {2015},
}

@article{lust_disintegration_2024,
    title = {Disintegration at the {Syntax}-{Semantics} {Interface} in {Prodromal} {Alzheimer}’s {Disease}: {New} {Evidence} from {Complex} {Sentence} {Anaphora} in {Amnestic} {Mild} {Cognitive} {Impairment} ({aMCI})},
    volume = {70},
    issn = {0911-6044},
    shorttitle = {Disintegration at the {Syntax}-{Semantics} {Interface} in {Prodromal} {Alzheimer}’s {Disease}},
    url = {https://pmc.ncbi.nlm.nih.gov/articles/PMC10871704/},
    doi = {10.1016/j.jneuroling.2023.101190},
    urldate = {2026-07-25},
    journal = {Journal of neurolinguistics},
    author = {Lust, Barbara and Flynn, Suzanne and Henderson, Charles and Gair, James and Sherman, Janet Cohen},
    month = may,
    year = {2024},
    pages = {101190},
}

@article{slegers_connected_2018,
    title = {Connected {Speech} {Features} from {Picture} {Description} in {Alzheimer}'s {Disease}: {A} {Systematic} {Review}},
    volume = {65},
    issn = {1875-8908},
    shorttitle = {Connected {Speech} {Features} from {Picture} {Description} in {Alzheimer}'s {Disease}},
    doi = {10.3233/JAD-170881},
    language = {eng},
    number = {2},
    journal = {Journal of Alzheimer's disease: JAD},
    author = {Slegers, Antoine and Filiou, Renée-Pier and Montembeault, Maxime and Brambati, Simona Maria},
    year = {2018},
    pages = {519--542},
}

@inproceedings{sirts_idea_2017,
    address = {Vancouver, Canada},
    title = {Idea density for predicting {Alzheimer}'s disease from transcribed speech},
    url = {https://aclanthology.org/K17-1033},
    doi = {10.18653/v1/K17-1033},
    urldate = {2022-04-29},
    booktitle = {Proceedings of the 21st {Conference} on {Computational} {Natural} {Language} {Learning} ({CoNLL} 2017)},
    publisher = {Association for Computational Linguistics},
    author = {Sirts, Kairit and Piguet, Olivier and Johnson, Mark},
    year = {2017},
    pages = {322--332},
}

@article{boschi_connected_2017,
    title = {Connected {Speech} in {Neurodegenerative} {Language} {Disorders}: {A} {Review}},
    volume = {8},
    issn = {1664-1078},
    shorttitle = {Connected {Speech} in {Neurodegenerative} {Language} {Disorders}},
    url = {https://pmc.ncbi.nlm.nih.gov/articles/PMC5337522/},
    doi = {10.3389/fpsyg.2017.00269},
    urldate = {2026-07-25},
    journal = {Frontiers in Psychology},
    author = {Boschi, Veronica and Catricalà, Eleonora and Consonni, Monica and Chesi, Cristiano and Moro, Andrea and Cappa, Stefano F.},
    month = mar,
    year = {2017},
    pages = {269},
}

@article{ivanova_defying_2023,
    title = {Defying syntactic preservation in {Alzheimer}'s disease: what type of impairment predicts syntactic change in dementia (if it does) and why?},
    volume = {2},
    issn = {2813-4605},
    shorttitle = {Defying syntactic preservation in {Alzheimer}'s disease},
    url = {https://www.frontiersin.org/journals/language-sciences/articles/10.3389/flang.2023.1199107/full},
    doi = {10.3389/flang.2023.1199107},
    language = {English},
    urldate = {2026-07-25},
    journal = {Frontiers in Language Sciences},
    publisher = {Frontiers},
    author = {Ivanova, Olga and Martínez-Nicolás, Israel and García-Piñuela, Elena and Meilán, Juan José G.},
    month = aug,
    year = {2023},
}

@article{kave_word_2018,
    title = {Word retrieval in connected speech in {Alzheimer}’s disease: a review with meta-analyses},
    volume = {32},
    issn = {0268-7038},
    shorttitle = {Word retrieval in connected speech in {Alzheimer}’s disease},
    url = {https://doi.org/10.1080/02687038.2017.1338663},
    doi = {10.1080/02687038.2017.1338663},
    number = {1},
    urldate = {2026-07-25},
    journal = {Aphasiology},
    publisher = {Routledge},
    author = {Kavé, Gitit and Goral, Mira},
    month = jan,
    year = {2018},
    note = {\_eprint: https://doi.org/10.1080/02687038.2017.1338663},
    pages = {4--26},
}

@misc{wang_emergent_2025,
    title = {Emergent modularity in large language models: {Insights} from aphasia simulations},
    copyright = {© 2025, Posted by Cold Spring Harbor Laboratory. This pre-print is available under a Creative Commons License (Attribution 4.0 International), CC BY 4.0, as described at http://creativecommons.org/licenses/by/4.0/},
    shorttitle = {Emergent modularity in large language models},
    url = {https://www.biorxiv.org/content/10.1101/2025.02.22.639416v1},
    doi = {10.1101/2025.02.22.639416},
    language = {en},
    urldate = {2026-07-28},
    publisher = {bioRxiv},
    author = {Wang, Chengcheng and Fan, Zhiyu and Han, Zaizhu and Bi, Yanchao and Li, Jixing},
    month = feb,
    year = {2025},
    note = {Pages: 2025.02.22.639416
Section: New Results},
}

@article{fedorenko_language_2024,
    title = {Language is primarily a tool for communication rather than thought},
    volume = {630},
    copyright = {2024 Springer Nature Limited},
    issn = {1476-4687},
    url = {https://www.nature.com/articles/s41586-024-07522-w},
    doi = {10.1038/s41586-024-07522-w},
    language = {en},
    number = {8017},
    urldate = {2024-08-08},
    journal = {Nature},
    publisher = {Nature Publishing Group},
    author = {Fedorenko, Evelina and Piantadosi, Steven T. and Gibson, Edward A. F.},
    month = jun,
    year = {2024},
    pages = {575--586},
}

@book{shallice_neuropsychology_1988,
  title={From Neuropsychology to Mental Structure},
  author={Shallice, Tim},
  year={1988},
  publisher={Cambridge University Press},
  address={Cambridge}
}

@inproceedings{geva_transformer_2021,
    address = {Online and Punta Cana, Dominican Republic},
    title = {Transformer {Feed}-{Forward} {Layers} {Are} {Key}-{Value} {Memories}},
    url = {https://aclanthology.org/2021.emnlp-main.446/},
    doi = {10.18653/v1/2021.emnlp-main.446},
    urldate = {2026-07-29},
    booktitle = {Proceedings of the 2021 {Conference} on {Empirical} {Methods} in {Natural} {Language} {Processing}},
    publisher = {Association for Computational Linguistics},
    author = {Geva, Mor and Schuster, Roei and Berant, Jonathan and Levy, Omer},
    editor = {Moens, Marie-Francine and Huang, Xuanjing and Specia, Lucia and Yih, Scott Wen-tau},
    month = nov,
    year = {2021},
    pages = {5484--5495},
}

@inproceedings{de_cao_sparse_2022,
    address = {Abu Dhabi, United Arab Emirates (Hybrid)},
    title = {Sparse {Interventions} in {Language} {Models} with {Differentiable} {Masking}},
    url = {https://aclanthology.org/2022.blackboxnlp-1.2/},
    doi = {10.18653/v1/2022.blackboxnlp-1.2},
    urldate = {2026-07-29},
    booktitle = {Proceedings of the {Fifth} {BlackboxNLP} {Workshop} on {Analyzing} and {Interpreting} {Neural} {Networks} for {NLP}},
    publisher = {Association for Computational Linguistics},
    author = {De Cao, Nicola and Schmid, Leon and Hupkes, Dieuwke and Titov, Ivan},
    editor = {Bastings, Jasmijn and Belinkov, Yonatan and Elazar, Yanai and Hupkes, Dieuwke and Saphra, Naomi and Wiegreffe, Sarah},
    month = dec,
    year = {2022},
    pages = {16--27},
}

@article{meng_locating_2022,
    title = {Locating and {Editing} {Factual} {Associations} in {GPT}},
    volume = {35},
    url = {https://proceedings.neurips.cc/paper_files/paper/2022/hash/6f1d43d5a82a37e89b0665b33bf3a182-Abstract-Conference.html?utm_source=chatgpt.com},
    doi = {10.52202/068431-1262},
    language = {en},
    urldate = {2026-07-29},
    journal = {Advances in Neural Information Processing Systems},
    author = {Meng, Kevin and Bau, David and Andonian, Alex and Belinkov, Yonatan},
    month = dec,
    year = {2022},
    pages = {17359--17372},
}

@inproceedings{nie_mechanistic_2025,
    address = {Suzhou, China},
    title = {Mechanistic {Understanding} and {Mitigation} of {Language} {Confusion} in {English}-{Centric} {Large} {Language} {Models}},
    isbn = {979-8-89176-335-7},
    url = {https://aclanthology.org/2025.findings-emnlp.37/},
    doi = {10.18653/v1/2025.findings-emnlp.37},
    urldate = {2026-07-29},
    booktitle = {Findings of the {Association} for {Computational} {Linguistics}: {EMNLP} 2025},
    publisher = {Association for Computational Linguistics},
    author = {Nie, Ercong and Schmid, Helmut and Schuetze, Hinrich},
    editor = {Christodoulopoulos, Christos and Chakraborty, Tanmoy and Rose, Carolyn and Peng, Violet},
    month = nov,
    year = {2025},
    pages = {690--706},
}

\appendix

\section{ADReSSo dataset}
\label{sec:appendix_adresso}
We used the Alzheimer's Dementia Recognition through Spontaneous Speech only (ADReSSo) dataset to identify neurons showing differential activation patterns between language produced by individuals with Alzheimer's disease and cognitively healthy controls \citep{luz_detecting_2021}. ADReSSo was developed as a benchmark for speech-based Alzheimer's disease recognition and was constructed from English-language recordings available through DementiaBank\footnote{https://talkbank.org/dementia/}. The dataset was designed to reduce common confounds in automated dementia recognition, including imbalances in age and sex and repeated observations from the same participant.

Each participant completed the Cookie Theft picture-description task from the Boston Diagnostic Aphasia Examination. Participants were asked to describe events in the picture in their own words, producing a short sample of spontaneous connected speech. The present study used the textual transcripts of the participant responses rather than the corresponding audio recordings.

The dataset contained 237 participants, comprising 122 individuals with probable AD and 115 cognitively healthy controls. Usage of this dataset aligns with the instruction of the DementiaBank ground rules. Each participant contributed one transcript.  Only speech produced by the participant was retained for the neuron-activation analysis. Interviewer utterances, transcription annotations, timing information, and other CHAT-format metadata were excluded. The resulting transcripts contained the lexical content of each participant's picture description. The demographic characteristics of each group are presented in Table~\ref{tab:adresso_demo}.

\begin{table}[tbhp]
\centering
\setlength{\tabcolsep}{2pt}
\begin{tabular}{lcc}
\hline
\textbf{Variables} & \textbf{Controls} & \textbf{Probable AD} \\
\hline
Participants ($n$) & 115 & 122 \\
Women ($n$) & 75 & 79 \\
Men ($n$) & 40 & 43 \\
Age (years) & $66.06 \pm 6.31$ & $69.38 \pm 6.88$ \\
Education (years) & $13.98 \pm 2.65$ & $11.97 \pm 2.63$ \\
Missing edu ($n$) & 22 & 0 \\
MMSE & $28.96 \pm 1.17$ & $17.84 \pm 5.48$ \\
Missing MMSE ($n$) & 1 & 0 \\
\hline
\end{tabular}
\caption{Characteristics of the ADReSSo sample used for neuron identification. Continuous variables are reported as mean $\pm$ standard deviation unless otherwise indicated.}
\label{tab:adresso_demo}
\end{table}

\section{Neuron distribution}
\label{sec:neuron_distribute}

We examined the layer-wise distribution of the neurons included in each intervention set. For each Transformer layer, we calculated the proportion of its 12288 feed-forward neurons that satisfied the corresponding selection criterion. The full set contained 67640 neurons with significantly higher activation rates for probable AD than for control transcripts after FDR correction. The top-2000 and top-10000 sets were obtained by globally ranking these eligible neurons according to the magnitude of their AD-oriented rank-biserial correlations. Experiments were conducted on the UPF high-performance computing cluster using four CPU cores for every job, each allocated 30 GB of memory. No model training or fine-tuning was performed.

\begin{figure*}[t]
    \centering
    \includegraphics[width=\linewidth]{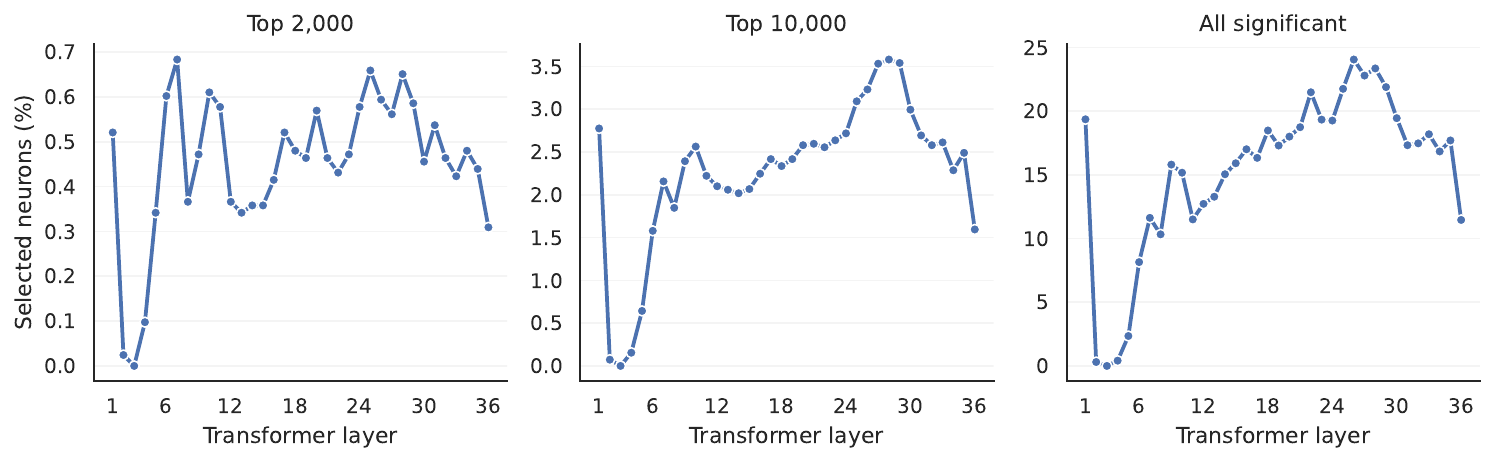}
    \caption{
    \textbf{Layer-wise distribution of AD-associated neurons across intervention scopes.}
    Each panel shows the percentage of the 12,288 feed-forward neurons in each Transformer layer included in the corresponding intervention set: the top 2000 neurons, the top 10000 neurons, or all 67640 significant AD-associated neurons. AD-associated neurons were defined as units with significantly higher activation rates for probable AD than for control transcripts after FDR correction ($q<0.05$). The top-ranked sets were selected globally according to the magnitude of the AD-oriented rank-biserial correlation. Percentages were calculated separately within each layer. The three panels use different vertical-axis scales.
    }
    \label{fig:neuron_distribution}
\end{figure*}

As shown in Figure~\ref{fig:neuron_distribution}, the selected neurons were distributed non-uniformly across the Transformer stack. Apart from the first layer, relatively few neurons were selected from the earliest layers, with no eligible neurons identified in layer 3. The proportion of selected neurons generally increased across the middle layers and was highest in the later-middle portion of the model, before declining toward the final layer. This broad layer-wise profile was visible across all three intervention scopes, although the smaller top-ranked sets showed greater local variation. These results indicate that the intervention targets were distributed across multiple layers rather than confined to a single localized component of the model.

\section{Prompts}
\label{sec:prompts}

\paragraph{Turn 0: fictional-participant initialization}

Hey, today we will complete some memory and thinking questions. You should role-play as one completely fictional \texttt{[ROLE]}, who has already retired\footnote{Every model performed 25 roles: farmer, welder, miner, cleaner, builder, nurse, doctor, truck driver, airport ground staff, emergency dispatcher, accountant, bank clerk, insurance agent, civil servant, salesperson, librarian, teacher, scientist, engineer, architect, police officer, soldier, firefighter, security guard, and prison officer.}. Firstly, you should choose random but plausible parameters consistent with that role, and state these parameters in one continuous paragraph: city, years of working, two personality traits, and current life focus. From then on, speak in the first person as this fictional participant. Answer naturally like a study participant. Some questions are easy, some are hard; nobody gets everything right. Please respond in one continuous paragraph (no bullet points), and keep it under 150 words. If you are not sure of an answer, say exactly “I don’t know” and stop. Do not add extra details you are unsure about.

\paragraph{Turn 1: immediate story recall}
I am going to read you a short story about a little bird. Please listen carefully. When I finish, tell me everything you can remember from the story in your own words. Try to include as many details as you can, but you may not remember everything. Do not quote the story word-for-word. Do not add new facts that were not in the story. If you are unsure about a detail, omit it. This is the story: "It was a hot day. A thirsty bird was looking for water for a long time. She was very tired and was about to faint. Suddenly, she spotted a pitcher of water under a bench in a park. She flew down and sat on the pitcher. She could see some water inside, but it was too deep for her to reach. Her beak was not long enough to drink the water. She looked around and found some pebbles. This gave her an idea. She picked up the pebbles one by one and dropped them into the pitcher. Soon the water level raised. She was now able to reach and drink it. The bird flew away happily." Respond in one continuous paragraph (no bullet points), and keep it under 150 words. Please begin your recall.

\paragraph{Turn 2: semantic animal fluency}
New task. I will give you a category. Please say as quickly as you can the names of things that belong to that category. The category is animals. List as many different animals as you can, but no more than 15 items. Stop after 15 items or when you cannot think of more. Only output animal names separated by commas, and nothing else. If you are not sure of an answer, say exactly “I don’t know” and stop. Do not add extra details you are unsure about. Please start. 

\paragraph{Turn 3: phonemic letter fluency}
New task. I will give you another category. Please say as quickly as you can the names of words that begins with C. List as many different words as you can, but no more than 15 items. Stop after 15 items or when you cannot think of more. Only output words separated by commas, and nothing else. If you are not sure of an answer, say exactly “I don’t know” and stop. Do not add extra details you are unsure about. Please start. 

\paragraph{Turn 4: digit span forward}
New task. I am going to say some numbers. Wait until I finish, and then repeat them back in the same order. Only write the numbers separated by spaces. Do not write any other words. Here are the numbers: 7 1 9 4 3 8 5. If you don’t know, say exactly “I don’t know.”

\paragraph{Turn 5: digit span backward}
New task. I am going to say some numbers again. This time I want you to repeat them in reverse order. Only write the numbers separated by spaces. Do not write any other words. Here are the numbers: 6 2 8 9 1 4 3. If you don’t know, say exactly “I don’t know.”

\paragraph{Turn 6: digit-letter span forward}
New task. I am going to say some numbers and letters. Wait until I finish, and then repeat them back in the same order. Only write the numbers and letters separated by spaces. Do not write any other words. Here are the items: 5 V 8 T M 4 6 N. If you don’t know, say exactly “I don’t know.”

\paragraph{Turn 7: digit-letter span backward}
New task. I am going to say some numbers and letters again. This time I want you to repeat them in reverse order. Only write the numbers and letters separated by spaces. Do not write any other words. Here are the items: 4 T B N 0 7 5 C. If you don’t know, say exactly “I don’t know.”

\paragraph{Turn 8: procedural knowledge}
New task. Describe how to make a cup of tea using a typical everyday method. Include the essential steps in the correct order. Do not add personal anecdotes. Do not include unnecessary or unsafe actions. Do not add details you are unsure about. Keep your answer under 150 words, in one paragraph, practical and clear, with no bullet points.

\paragraph{Turn 9: scene construction}
New task. Imagine you are shopping in a supermarket. Describe the scene as if you were there. Do not recount an actual memory, but construct a new everyday scene. Keep it realistic. If you don’t know, say exactly “I don’t know.” Do not add details you are unsure about. Keep your answer under 150 words, in one paragraph, with no bullet points.

\paragraph{Turn 10: coreference resolution}
Great job! New task. Rewrite the text so that a reader will never be unsure what each pronoun refers to. Replace all pronoun with the appropriate name or noun phrase and no pronoun should remain in the rewritten text. After rewriting, verify that no pronouns remain. Do not change the order of events, add new events, or introduce new information. If you can’t complete the task, say exactly “I don’t know.” Keep the same number of sentences if possible. Text: "Horace was crossing a quiet square when a small cat jumped onto a fountain and knocked a shiny coin into the water. He reached down to retrieve it, but the cat splashed the water and darted away with a flick of its tail. This startled several pigeons nearby, and they scattered across the square as he laughed at the sudden chaos. A moment later, it returned and watched him from a safe distance." 

\paragraph{Turn 11: delayed memory recall}
Great, thank you. The final task. A few turns ago, I read you a story about a little bird. Now please tell me everything you can remember about that story, in your own words. Try to include as many details as you can, but do not quote the story word-for-word. Do not add new facts that were not in the story. Do not add new facts that were not in the story. If you are unsure about a detail, omit it.  Respond in one continuous paragraph (no bullet points), and keep it under 150 words. Please begin your recall. 

\section{Rating scheme}
\label{sec:rating_scheme}

The raters annotated the responses using Label Studio\footnote{https://labelstud.io/}, although the raters were not necessarily the same across tasks. All raters were students and had completed at least undergraduate-level training in linguistics. They were all proficient in English. The two raters responsible for the more complex task, episodicity annotation, were doctoral candidates in linguistics. The first author conducted a general quality check of the annotations but did not replace the independent ratings. All raters were paid for the annotations. 

\subsection{Story recall}
From the story, we identified seven entities and twelve events. Raters were instructed to label if any of them is missing in the recall. 

Entities are: (1) bird, (2) water, (3) pitcher, (4) pebbles, (5) bench, (6) park, and (7) beak (or beak length). 

Events are: (1) The day is hot. (2) The bird is thirsty. (3) The bird looks for water for a long time. (4) The bird becomes very tired and is about to faint. (5) The bird spots a pitcher of water. (6) The pitcher is under a bench in a park. (7) The bird flies down and sits on the pitcher. (8) The water is too deep to reach, the bird's beak is not long enough. (9) The bird finds some pebbles. (10) The bird picks up pebbles and drops them into the pitcher. (11) The water level rises, allowing the bird to drink. (12) The bird flies away happily.

\subsection{Procedural discourse}
We identified six essential steps for making a cup of tea: (1) Heat water (kettle/pot; heat to hot/boiling). (2) Add tea (teabag or loose tea into cup/mug/teapot). (3) Pour hot water (water added to tea; tea is submerged). (4) steep / wait (leave to infuse; e.g., 3–5 minutes). (5) End steeping \& separate tea from water (remove bag / strain leaves). (6) Completion \& optional adjustments (stir; milk/sugar/lemon/honey optional; drink)

Raters also examined the organization of the discourse. The response got 0 if it is totally disorganized with no executable order or major reversals, 1 if it is partially ordered but messy with some reversals or back-and-forth, and 2 if it is coherent and logically ordered without major reversals but allowing some minor gaps. 

\subsection{Scene construction}
Following \citet{he_episodic_2024}, raters classified every sentence and embedded clause as episodic (EP), non-episodic (NONEP), and other meta statements (OTHER, e.g., I don't know). 

\subsection{Coreference resolution}
There are eight pronouns to be resolved in the original texts, which are highlighted with brackets:
\begin{quote}
    Horace was crossing a quiet square when a small cat jumped onto a fountain and knocked a shiny coin into the water. [He] reached down to retrieve [it], but the cat splashed the water and darted away with a flick of [its] tail. [This] startled several pigeons nearby, and [they] scattered across the square as [he] laughed at the sudden chaos. A moment later, [it] returned and watched [him] from a safe distance.
\end{quote}

Gold answers: (1) he: Horace; (2) it: shiny coin; (3) its: the small cat's; (4) this: the cat splashing the water and darting away / the sudden chaos; (5) they: several pigeons nearby; (6) he: Horace; (7) it: the small cat; (8) him: Horace. 

\begin{figure*}
    \centering
    \includegraphics[width=\linewidth]{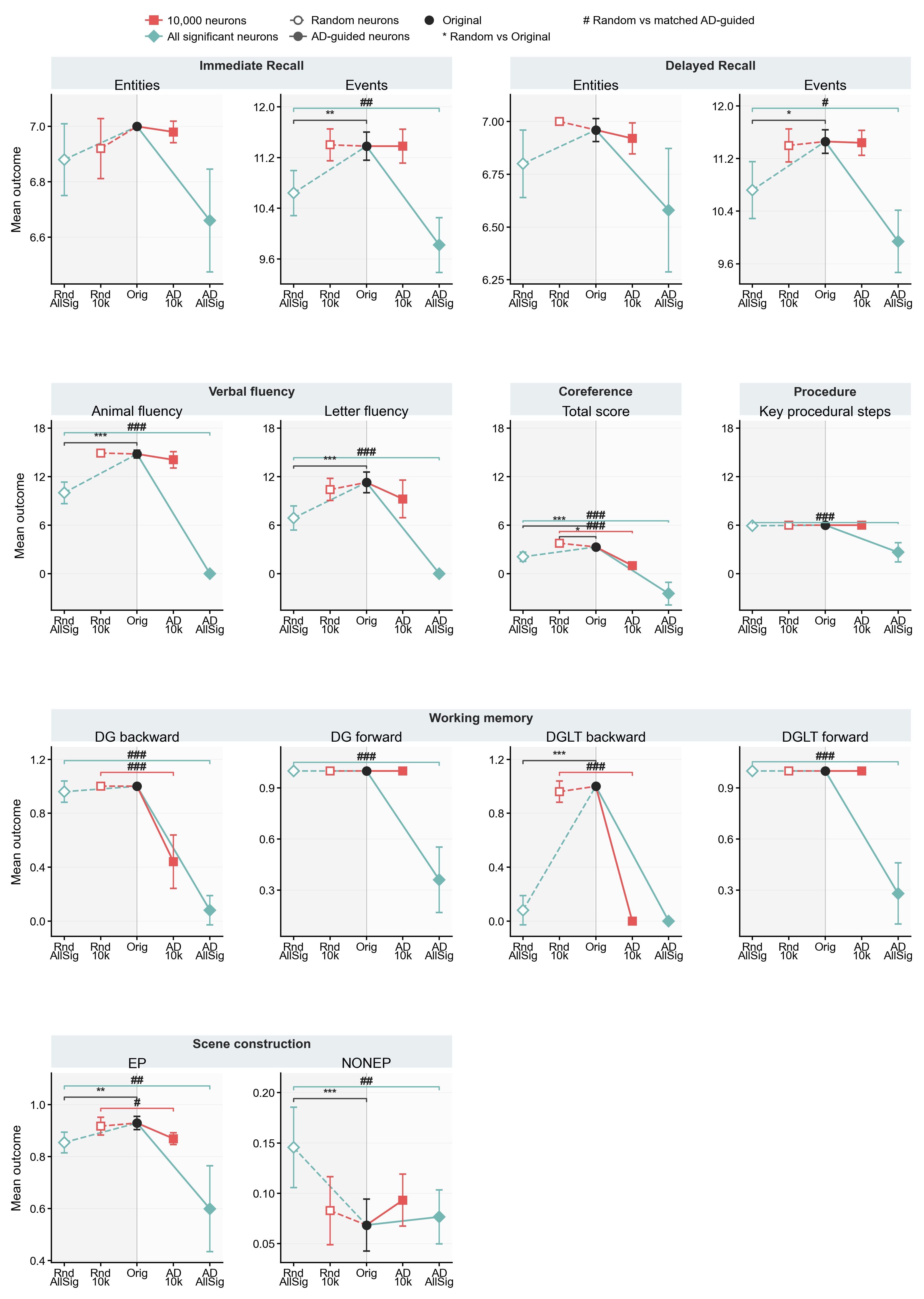}
    \caption{
    \textbf{Effects of AD-guided neuron intervention across cognitive–linguistic tasks.}
    Error bars indicate 95\% confidence intervals. Asterisks denote comparisons between a random-neuron condition (Rnd) and the original model (Orig), whereas hash symbols denote comparisons between random-neuron (Rnd) and size-matched AD-guided interventions (AD). DG: digit; DGLT: digit-letter; EP: episodic; NONEP: non-episodic. Asterisks: $^{*}q<0.05$, $^{**}q<0.01$, $^{***}q<0.001$.
    }
    \label{fig:random_control}
\end{figure*}

\section{Random neuron intervention control}
\label{sec:random_control}

\paragraph{Design and analysis.}
We conducted a post-hoc random neuron control analysis to examine whether the observed performance changes could be attributed to nonspecific perturbation from editing many feed-forward neurons. For each Transformer layer, we randomly sampled the same number of neurons as in the corresponding conditions of AD-associated top 10,000 neurons and of all AD-associated neurons. The random controls therefore matched both the total number and the layer-wise distribution of edited neurons. These neurons were amplified using the same strongest intervention strength ($\alpha=0.6$), which resulted in the most obvious changes in the main experiment, while the prompts, occupational profiles, generation settings, and assessment procedures remained unchanged.

The random conditions were compared with both the original model and the corresponding AD-guided condition using the same GEE framework as in the main analysis. Four planned contrasts were examined within each outcome: random 10,000 versus original, random all significant versus original, random 10,000 versus AD-guided 10,000, and random all significant versus AD-associated all significant. The four contrasts were FDR-corrected separately within each outcome.

\paragraph{Annotation limitation.}
This control was added after completion of the primary experiment and was not part of the original annotation plan. The original raters were no longer available to evaluate the additional outputs. Responses requiring manual assessment were therefore annotated by the first author using the same task-specific scoring criteria as in the primary analysis. Accordingly, this analysis should be interpreted as a post hoc sensitivity and validation control, rather than as an independently rated replication of the primary results.

\paragraph{Results.}
As shown in Figure~\ref{fig:random_control}, randomly editing 10,000 neurons generally produced smaller changes than editing the corresponding AD-associated neurons. Performance in the random 10,000-neuron condition remained close to the original model for most outcomes, whereas the AD-associated top 10,000 neuron condition showed clearer reductions, particularly in coreference resolution and backward working-memory tasks. This pattern suggests that these effects were not explained solely by modifying an equivalent number of arbitrary neurons.

The layer-matched random all-significant intervention produced broader changes, including reductions in verbal fluency, coreference resolution, some working-memory outcomes, and scene-construction performance. Thus, perturbing a sufficiently large number of neurons can itself produce nonspecific performance degradation. Nevertheless, the AD-guided all-significant condition frequently showed larger or distinct changes from the random intervention, particularly for verbal fluency, coreference resolution, procedural discourse, working memory, and episodic detail production. Entity recall was comparatively stable, whereas event recall showed greater sensitivity to random perturbation. In scene construction, both two randomly edited models did not produce any OTHER statements, as the original QWen3-8B model. 

Given that episodic memory decline is a hallmark of AD, the random interventions' failure to consistently reproduce the same impairment profile across immediate recall, delayed recall, and scene construction provides further, although preliminary, support for the specificity of AD-guided neuron selection.

\section{Automated linguistic measures}
\label{sec:nlp_feature}
\paragraph{Preprocessing.} Responses were processed using the \texttt{en\_core\_web\_lg} pipeline in spaCy. Tokens tagged as punctuation, spaces, symbols, or unclassified elements were excluded from the analyses. All remaining tokens were converted to lowercase.

\paragraph{Response length.} Token count was defined as the number of retained tokens in each response. 

\paragraph{Lexical richness.} Moving-average type-token ratio was calculated over successive 25-token windows, with the type-token ratio calculated separately in each window and then averaged across the response. For responses containing 25 tokens or fewer, the type-token ratio of the complete response was used. Lower values indicated greater lexical repetition and more restricted vocabulary. 

\paragraph{Lexical density.} Lexical density was calculated as $N_{\mathrm{content}}/N_{\mathrm{tokens}}$, where content words comprised nouns, verbs, adjectives, and adverbs.

\paragraph{Propositional density.} DEPID-R was used to quantify non-redundant propositional information \citep{sirts_idea_2017}. Dependency relations corresponding to propositions were represented as tuples containing the dependency relation, dependent lemma, and head lemma. Definite and indefinite articles and non-referential subject uses of \textit{it} and \textit{this} were excluded \citep{sirts_idea_2017}. Repeated tuples were counted only once, and DEPID-R was calculated as the number of unique propositions divided by the number of retained tokens. 

\paragraph{Clause ratio.} Clause ratio was calculated as the number of tokens assigned clause-related dependency relations divided by the total number of non-punctuation dependency tokens. The included relations comprised clausal subjects, clausal complements, open clausal complements, adverbial clauses, and adnominal or relative-clause modifiers. Higher values indicated a greater density of clausal elaboration and indexed greater hierarchical syntactic complexity. 

\paragraph{Lexical surprisal.} Frequency-based lexical surprisal was calculated for each token as $-\log_2 p(w)$, where $p(w)$ was its English unigram frequency obtained from the \texttt{wordfreq} database\footnote{https://github.com/rspeer/wordfreq}. Values were averaged across all tokens in the response, with a minimum probability of $10^{-9}$ assigned to words without an available frequency estimate. Higher values indicated the use of less frequent vocabulary. 

\end{document}